\documentclass[11pt]{article}
\usepackage{acl}

\usepackage{times}
\usepackage[T1]{fontenc}
\usepackage[utf8]{inputenc}
\usepackage{microtype}
\usepackage{inconsolata}
\usepackage{graphicx}
\usepackage{subcaption}

\usepackage{comment}
\usepackage{footmisc}
\usepackage{tabularx}
\usepackage{booktabs}
\usepackage{xcolor, soul}

\title{How Architecture, Typology, and Adaptation Strategies Shape ASR for Low-Resource African Languages: A Controlled Empirical Study}

\author{
    \begin{minipage}{0.9\textwidth} \centering \normalfont
    Mercy Muchai$^{1}$,
    Kevin Chege$^{1}$,
    Nick Mumero Mwangi$^{1}$,
    Samuel Chege Maina$^{1}$,
    Felermino Ali$^{1}$,
    Elizabeth Ankrah$^{1}$,
    Ignatius Ezeani$^{2}$ \\
    $^{1}$Microsoft Research Africa, Nairobi \\
    $^{2}$Lancaster University
    \end{minipage}
}

\begin{document}

\maketitle

\begin{abstract}
Despite recent advancements in ASR, identifying effective architectures, fine-tuning approaches, and data adaptation strategies for low-resource languages remains challenging. We present a systematic comparison across five low-resource African languages and Swahili as a mid-resource contrast, fine-tuning three models from distinct architectures: Phi-4-multimodal-instruct, Whisper-large-v3-turbo, and MMS-1B-all. We analyze data sufficiency, monolingual versus multilingual adaptation, domain fine-tuning, and language-adapter transfer. Phi-4 achieves the best overall accuracy, MMS shows comparable CER, while Whisper exhibits high CER, consistent with known hallucinations and repetition loops. Performance saturates at different data scales, offering practical data-efficiency targets, while typological complexity consistently shapes error rates. Domain fine-tuning benefits Phi-4 and Whisper but degrades MMS, revealing limitations of adapter-based adaptation. Proxy language adapters outperform training MMS adapters from scratch.
\end{abstract}

\section{Introduction}
In low-resource ASR, data scarcity cannot be addressed by scaling labeled datasets alone. Researchers must make careful choices about pretrained models, training strategies, data allocation, and domain-specific supervision. These choices determine not only model performance but also how effectively limited collection and annotation resources are used. Improving this process is essential for extending speech technologies beyond high-resource languages and ensuring that linguistically diverse communities are not excluded from advances in voice-based systems.

Advances in automatic speech recognition (ASR) have approached human-level performance on selected high-resource benchmarks. However, many of the world’s languages still lack the hundreds or thousands of hours of transcribed speech commonly used to develop high-performing systems \citep{LiuYangQu2024WhisperFinetuning,LiangLevow2025TranscriptionBottleneck}. This scarcity contributes to substantially higher recognition error rates and limits the availability of reliable speech technologies for speakers of under-represented languages.

Pretrained multilingual ASR models offer a promising way to compensate for limited labeled data by transferring knowledge learned from other languages and domains \citep{pratap2024scaling}. Models such as Whisper \citep{radford2023whisper}, Phi-4 \citep{agrawal2025phi4multimodal}, and MMS \citep{pratap2024scaling} provide strong starting points, but their effectiveness depends on several design decisions. Practitioners must determine which model and architecture is most suitable for a given data budget, how much labeled speech is needed to achieve useful performance, whether languages should be fine-tuned independently or jointly, how models should be initialized for unsupported languages, and whether scarce annotation effort should target general or domain-specific speech. In low-resource settings, unsuitable choices can waste a substantial proportion of the available data and computational resources.

This paper investigates these decisions through a systematic series of experiments on six African languages. We fine-tune three models from distinct architectures under controlled per-language data budgets and evaluate their performance using character error rate (CER) and word error rate (WER) metrics. Our experiments examine data sufficiency and diminishing returns, compare model architectures and fine-tuning strategies, evaluate proxy-language adapter transfer effectiveness versus new adapter initialization, and investigate the role of domain adaptation.  Our contributions include: 



\begin{itemize}
    \item An empirical characterization of the relative performance of three pretrained ASR systems across matched per-language data budgets and low-resource conditions.

    \item Evidence on data sufficiency in low-resource ASR, including the data ranges associated with larger performance gains and diminishing returns.

    \item Evidence on low-resource training strategies, comparing monolingual and multilingual fine-tuning and proxy-language versus new adapter initialization.

    \item Evidence on domain adaptability in low-resource ASR, examining how domain-specific training data affects in-domain performance and generalization to unseen data from the same domain.
\end{itemize}

The evaluated systems differ in parameter count, pretraining data, trainable components, and adaptation objectives, with each architecture represented by one model. We therefore interpret the results as model-level comparisons rather than isolated causal effects of architecture. 

Rather than seeking a universally optimal system, we identify how these design choices perform across different low-resource conditions. Collectively, these findings provide practical evidence for collecting speech data and training ASR systems when resources are limited.

\section{Related Work}

Recent studies have attempted to quantify the amount of transcribed speech required to build effective automatic speech recognition (ASR) systems for low-resource languages. Evidence suggests that ASR performance improves rapidly with the first few hours of labeled speech, after which gains gradually plateau as additional data are added \citep{bartelds-etal-2023-making}. However, the exact data requirements vary considerably across studies. Some studies report ASR systems trained with only a few minutes to a few hours of transcribed speech, particularly when the limited labeled data are supplemented with unlabeled or synthetic speech through data augmentation techniques \citep{bartelds-etal-2023-making,11385096,mainzinger-levow-2024-fine}.

To further reduce the need for large annotated corpora, several approaches have been proposed. Self-training enables models to generate pseudo-labels for unlabeled speech \citep{bartelds-etal-2023-making}, while cross-lingual representation learning transfers knowledge from multilingual speech data to low-resource languages \citep{babu2022xlsr}, significantly reducing the transcription burden in low-resource settings.


Multilingual fine-tuning has been explored as a strategy to improve sample efficiency through shared cross-language learning, but may also cause interference and per-language degradation \citep{babu2022xlsr}. However, many evaluations increase the total training data alongside the number of languages, confounding the benefits of multilingual adaptation. Controlled comparisons that keep per-language training hours constant remain rare, limiting practical guidance on minimum labeled data requirements and the conditions under which multilingual transfer is beneficial in low-resource ASR.

Additionally, ASR models for low-resource languages demonstrate substantial performance degradation under domain mismatch, with error rates increasing sharply when evaluated outside their training domains \citep{li2021domain}. Both hybrid and sequence-to-sequence architectures benefit from limited domain supervision, where one to two hours of in-domain audio or text can yield large relative error reductions \citep{soky2022domain,mak2024exploring}. While multilingual self-supervised pretraining reduces data requirements, it does not eliminate sensitivity to domain-specific acoustic and lexical variation, underscoring the need for evaluations that explicitly account for cross-domain robustness \citep{dalmia2018domain}.

Parameter-efficient adaptation methods such as adapters \cite{houlsby2019parameter}, enable language-specific specialization atop a shared multilingual backbone. While promising for related languages with shared phonetic structure, empirical evidence for adapter-based transfer in low-resource African languages remains sparse.

\section{Experimental Setup}
To address our research questions, we conduct a series of experiments that evaluate and compare performance across variations in data sufficiency, domain robustness, and fine-tuning strategies for low-resource languages. 

\subsection{Dataset Selection}
We select six diverse African languages: Swahili \texttt{(swa)}, a mid-resource language, and five low-resource languages: Dholuo \texttt{(luo)}, Kalenjin \texttt{(kln)}, Kikuyu \texttt{(kik)}, Maasai \texttt{(mas)}, and Somali \texttt{(som)}. The first four are predominantly spoken in Kenya, while Somali is spoken across the Horn of Africa. Together the six span three language families (Table~\ref{tab:language_families}), providing typological diversity for our analysis. According to Ethnologue \citep{ethnologue2024}, each language is classified by family with its corresponding ISO 639-3 code.

\begin{table}[h]
\centering
\begin{tabular}{l c r}
\toprule
\textbf{Language Family} & \textbf{Lang Code} & \textbf{Hours} \\
\midrule
Nilo-Saharan & \texttt{luo} & 513.2 \\
              & \texttt{kln} & 386.7 \\
              & \texttt{mas} & 235.5 \\
\addlinespace
Niger-Congo   & \texttt{kik} & 967.8 \\
              & \texttt{swa} & 280.0 \\
\addlinespace
Afro-Asiatic  & \texttt{som} & 341.9 \\
\bottomrule
\end{tabular}
\caption{Language families, ISO 639-3 codes, and available speech hours from the listed corpora}
\label{tab:language_families}
\end{table}

We curate the \texttt{luo, kik, kln, mas}, and \texttt{som} datasets from the Africa Next Voices–Kenya (ANV-KE) corpus\footnote{\label{fn:anv_data_ke}\url{https://huggingface.co/datasets/MCAA1-MSU/anv_data_ke}}. We further augment the \texttt{kik} corpus with agricultural-domain speech data\footnote{\label{fn:dg_green_kikuyu}\url{https://huggingface.co/datasets/DigiGreen/KikuyuASR_trainingdataset}}. The \texttt{swa} dataset is sourced from Mozilla Common Voice\footnote{\url{https://datacollective.mozillafoundation.org/datasets/cmj8u3puq00qhnxxbg26y0owu}}.

In addition, we evaluate models on the test split of a private agricultural Kikuyu dataset for domain robustness\footnote{\textit{\label{fn:kik_private} Private Kikuyu (\texttt{kik}) speech dataset in the agricultural domain, consisting of manually transcribed read speech, used only for evaluation}}. This non-public dataset is included to assess model generalization to domain-specific, real-world speech that is underrepresented in publicly available corpora, and has no overlap with training data. The ANV-KE dataset spans 11 content domains across the five languages, ensuring exposure to diverse vocabulary while the Swahili (\texttt{swa}) dataset from Mozilla Common Voice \citep{ardila2020commonvoice} does not include domain annotations.

\subsection{Dataset Preprocessing}
All audio data are resampled to 16 kHz prior to training and evaluation. Text normalization consists of lowercasing, collapsing repeated whitespace, and removing punctuation, while retaining language-specific diacritics. Evaluation is performed on the original test splits for each language dataset, whose sizes remain unchanged. WER and CER are length-normalized, enabling comparison across test sets of different sizes.

\subsection{Model Selection}
We select three models, each from a distinct architecture based on how speech is encoded, how text is generated, and whether decoding is autoregressive or non-autoregressive. Each model has varying language support and parameter count as shown in Table~\ref{tab:model_overview}. For the encoder-decoder architecture, we select \texttt{Whisper} (\texttt{openai/whisper-large-v3-turbo}) \citep{radford2023whisper}, which supports 99 languages. For the encoder-only architecture, we select \texttt{MMS} (facebook/mms-1b-all) \citep{pratap2024scaling} which was built on top of the \texttt{Wav2Vec2.0} \citep{Baevski2020} architecture with language coverage of 1,162 languages. For the LLM-multimodal architecture, we select \texttt{Phi-4} (\texttt{microsoft/Phi-4-multimodal-instruct}) \citep{agrawal2025phi4multimodal} which reports ASR support for 8 languages and was, at the time of our experiments, the highest-ranked multilingual LLM-based ASR model on the OpenASR leaderboard \citep{srivastav2025openasrleaderboardreproducible}. 
Together, these models cover distinct ASR model architectures and multilingual scales, enabling a comparison of performance, adaptability, and computational cost across model families.
\begin{table}[h]
\centering
\small
\setlength{\tabcolsep}{3pt}
\begin{tabular}{@{}l r r p{2.8cm}@{}}
\toprule
\textbf{Model} &
\shortstack{\textbf{Speech}\\\textbf{Params}} &
\shortstack{\textbf{Languages}\\\textbf{Supported}} &
\textbf{Architecture} \\
\midrule
Phi-4   & 920\,M\footnotemark & 8     & LLM multimodal \\
Whisper & 809\,M              & 99    & Encoder--decoder \\
MMS     & 1.0\,B              & 1,162 & Encoder-only \\
\bottomrule
\end{tabular}
\caption{Overview of the three ASR models evaluated in this study.}
\label{tab:model_overview}
\end{table}

\footnotetext{For Phi-4, the speech-specific parameters comprise the audio encoder and modality projector (460M) together with the speech LoRA (460M) \citep{agrawal2025phi4multimodal}, while the underlying 5.6B language model is shared across modalities.}
For Whisper, we extend the tokenizer during fine-tuning to include the language tokens and resize the input embeddings accordingly. \texttt{Swahili (swa)} is excluded, as it is already natively supported. For the Phi-4 model, only the audio-specific components: audio embedding module, audio encoder, and audio projection layers, are unfrozen and set as trainable, while the rest of the model parameters remain frozen to preserve pretrained language capabilities. For the MMS model, we fine-tune existing language-specific adapter layers while keeping the pretrained MMS backbone frozen. For languages not originally supported by MMS, \texttt{Kalenjin (kln)} and \texttt{Maasai (mas)}, we initialize from adapters of closely related languages; see Section~\ref{sec:adapters}), selected by genealogical proximity within the Kalenjin and Maa subgroups.

\subsection{Hardware and Hyperparameters}
\label{sec:hardware}

\textbf{Hardware}: All experiments are conducted using distributed data-parallel (DDP) training. MMS and Whisper models are trained on 8× NVIDIA A100 GPUs (40GB), while the Phi model is trained on 4× NVIDIA A100 GPUs (80GB).

\textbf{Hyperparameters}: All experiments follow the default fine-tuning hyperparameter configurations recommended in each of the model publications and official fine-tuning documentation for each model: MMS \citep{pratap2020massively}, Whisper \citep{radford2023whisper}, and Phi-4 \citep{agrawal2025phi4multimodal}. Optimizer settings, learning rate schedules, weight decay, and other regularization parameters are held constant within each model across all experiments to ensure comparability. The complete set of fine-tuning hyperparameters is summarized in Table~\ref{tab:training_hparams}.

\subsection{Fine-tuning Strategies}

\subsubsection{Multilingual vs Monolingual}
We evaluate multilingual versus monolingual model fine-tuning. For monolingual training, we fine-tune each model on one language at a time, while for multilingual fine-tuning, we create a unified multilingual dataset comprising all six languages, and perform joint fine-tuning, alternating batches across languages to promote cross-lingual transfer. Because MMS routes each language through a dedicated adapter, joint multilingual fine-tuning is not applicable; multilingual experiments therefore cover Phi-4 and Whisper only.

\subsubsection{New vs Proxy Adapter Training}
\label{sec:adapters}
We compare two language-adapter fine-tuning strategies for low-resource languages specific to the MMS model: \textit{(i) training adapters from scratch} versus \textit{(2) fine-tuning adapters from closely related languages}. Experiments are conducted on \texttt{mas} and \texttt{kln} that lack adapters in the base model. We select adapters from Samburu (\texttt{saq}) for Maasai and Sabaot (\texttt{spy}) for Kalenjin, based on genealogical proximity (Samburu is a Maa language mutually intelligible with Maasai; Sabaot is a Kalenjin variety). For each language, both strategies are fine-tuned on the same data and evaluated on identical held-out sets.

\subsection{Data Sufficiency}
To analyze the impact of data scaling, we fine-tune all three ASR models on up to 200 hours of speech per language. For MMS and Phi-4, we extend training to \(\approx\)450 hours to probe for performance saturation; Whisper is capped at 200\,h due to the substantially higher CPU overhead of its log-Mel feature extraction pipeline at larger data scales.  Model checkpoints are saved progressively after approximately \textit{5, 10, 20, 30, 40, 50, 75, 100, 150, 200, 300, 400 and 450} hours of cumulative exposure. The denser sampling in the early stages captures steep adaptation dynamics commonly observed in low-data regimes, while wider intervals at higher scales allow us to track longer-term performance trends and potential saturation effects, consistent with general findings from neural scaling studies \citep{li2021scaling}.
For Phi-4 and Whisper, fine-tuning is conducted in a multilingual setting using a unified dataset comprising all six target languages. In contrast, MMS is fine-tuned independently for each language using language-specific adapters, leveraging MMS’s language-specific adapter framework. Each checkpoint is evaluated separately for all six target languages using WER and CER.

\subsection{Domain Adaptability} \label{sec:domain_adaptability}
To evaluate domain adaptability, we fine-tune and evaluate each model under these conditions:
\begin{itemize}
\item \textbf{Cross-Domain Only}: the model is fine-tuned on 200-hour  general-domain Kikuyu data from the ANV-KE corpus spanning 11 content domains. This measures zero-shot transfer to the agriculture domain.
\item \textbf{Agriculture Domain (in-domain)}: the Cross-Domain training set is augmented with 38.3 h of in-domain agricultural speech\protect\footref{fn:dg_green_kikuyu} ($\approx$19\% of the 200 h base), and the model is evaluated on a held-out split of that same agricultural corpus. This measures in-domain adaptation.
\item \textbf{Agriculture Domain (out-of-sample)}: the same agriculture-augmented model is evaluated on a separate, private agricultural Kikuyu test set drawn from a different source, with no overlap with training\protect\footref{fn:kik_private}. This measures generalization to unseen agricultural speech.
\end{itemize}

Because the augmented condition also adds training data, in-domain gains partly conflate domain match with data quantity; however, our data-sufficiency curves (Fig.~2) show that a 200→238 h increase yields <1 pp CER gain in in-domain-free conditions, so the large reductions observed here are attributable primarily to domain match.

\section{Results and Discussion}\label{sec:results}
We report results for all the above experiments comparing performance across the models while prioritizing CER, because it is less sensitive than WER to word-boundary ambiguity and errors in morphologically complex words \citep{k-etal-2025-advocating}. Our experiments yield principal findings with direct practical implications for ASR development in low-resource language settings.

\subsection{Model Architecture Comparison}
\begin{figure}[!t]
\centering
\includegraphics[width=\columnwidth]{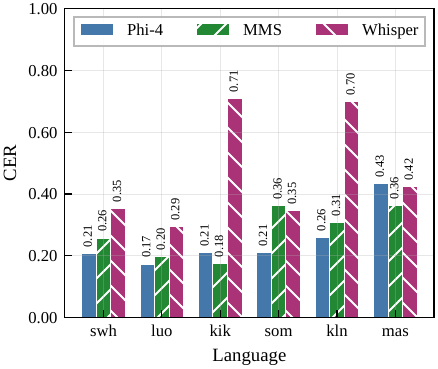}
\caption{CER across six languages for Phi-4, MMS, and Whisper fine-tuned on 200\,h of monolingual data. Corresponding WER is reported in Figure~\ref{fig:model_comparison_wer} (Appendix~\ref{sec:appendix_wer}).}
\label{fig:model_comparison_cer}
\end{figure}

\textbf{Architecture matters more than pre-training coverage.} Figure~\ref{fig:model_comparison_cer} presents the CER and WER distributions across the six languages for the three models.  Phi-4 achieves the best CER in four of the six languages, despite initial support for only eight languages, indicating strong adaptation capability against ASR-only models given its LLM contextual understanding. MMS demonstrates comparable CER performance to Phi-4, despite initial coverage of 1,100+ languages, ranking as the second-best model. In contrast, Whisper performs poorly on CER metrics, with errors compounded by its documented susceptibility to hallucinations from non-speech audio and repeated-text looping \citep{baranski2025whisper}. We do not apply any post-processing to mitigate Whisper's end of sequence hallucinations.

\subsection{Data Sufficiency}
CER and WER improve substantially with increased training data, but the trajectory beyond 150\,h (Table~\ref{tab:pct_change}) indicates that \textbf{data saturation is model-dependent.} To assess whether additional data yields a statistically significant CER change, we apply paired $t$-tests across the six languages for each model.

\begin{figure*}[!t]
\centering
\begin{subfigure}{0.49\textwidth}
\centering
\includegraphics[width=\linewidth]{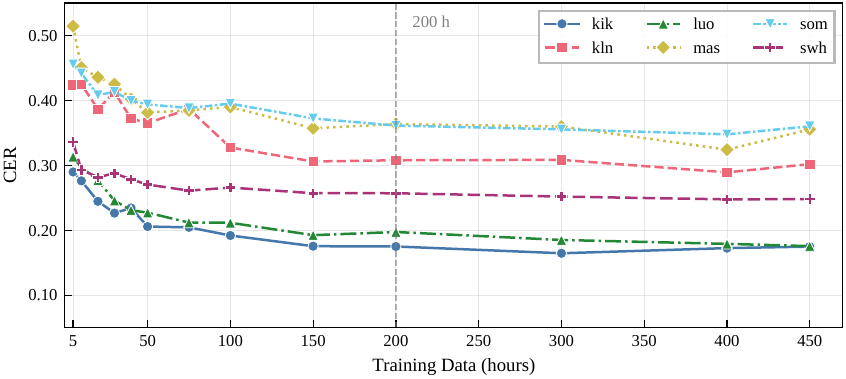}
\caption{MMS (monolingual), 5--450\,h.}
\label{fig:cer_data_sufficiency}
\end{subfigure}\hfill
\begin{subfigure}{0.49\textwidth}
\centering
\includegraphics[width=\linewidth]{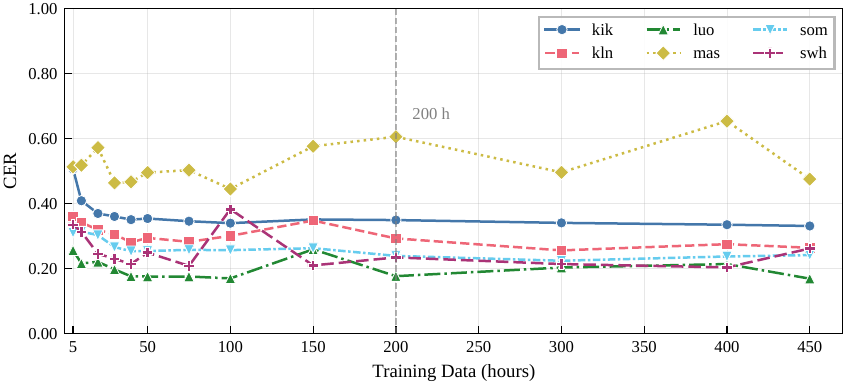}
\caption{Phi-4 (multilingual), 5--450\,h.}
\label{fig:phi_cer_data_sufficiency}
\end{subfigure}
\caption{CER as a function of training data (hours) across six languages. Performance largely plateaus after 150\,h.}
\label{fig:data_sufficiency}
\end{figure*}

MMS changes by only $-0.7$ percentage points (pp) from 150\,h to 450\,h ($p{=}0.036$), indicating practical saturation (Figure~\ref{fig:cer_data_sufficiency}). Phi-4 improves on five languages through 300\,h (mean $\Delta{=}{-}4.4$\,pp), but Swahili degrades by $+5$\,pp and the overall change is not significant ($p{=}0.126$; Figure~\ref{fig:phi_cer_data_sufficiency}). Whisper is excluded from this extended analysis because its log-Mel feature extraction pipeline introduces substantially higher CPU overhead at larger data scales, capping training at 200\,h. An initial 150--200\,h target is therefore reasonable for adapter-based models, while the optimal data regime for multilingual LLM-based models is language-dependent.

\subsection{Multilingual versus Monolingual}
\begin{figure*}[t]
\centering
\begin{subfigure}{0.49\textwidth}
\centering
\includegraphics[width=\linewidth]{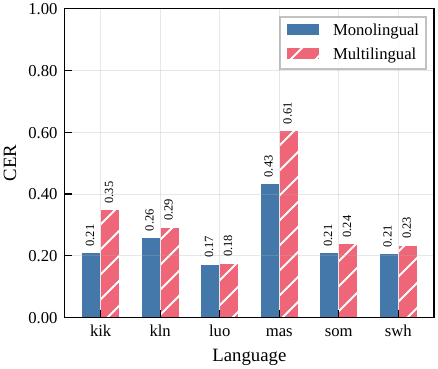}
\caption{Phi-4}
\label{fig:mono_vs_multi_phi}
\end{subfigure}\hfill
\begin{subfigure}{0.49\textwidth}
\centering
\includegraphics[width=\linewidth]{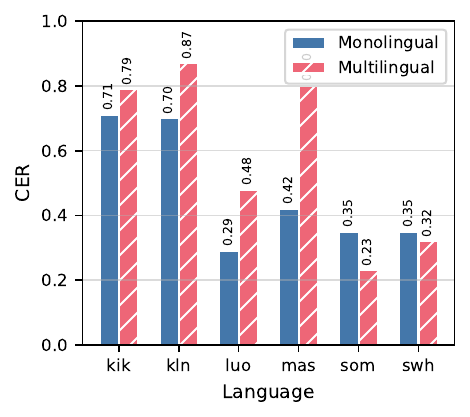}
\caption{Whisper}
\label{fig:mono_vs_multi_whisper}
\end{subfigure}
\caption{Monolingual vs.\ multilingual CER at 200\,h per language for Phi-4 (left) and Whisper (right).}
\label{fig:mono_vs_multi}
\end{figure*}

We evaluate multilingual joint fine-tuning for Phi-4 and Whisper, contrasting this with monolingual fine-tuning. Figures~\ref{fig:mono_vs_multi_phi} and~\ref{fig:mono_vs_multi_whisper} show that \textbf{multilingual training does not guarantee positive transfer.} Monolingual fine-tuning outperforms multilingual on 10 of 12 language-model pairs, though individual exceptions exist (e.g., Whisper Somali CER improves from 0.35 mono to 0.23 multi). For multilingual models, training on the unified dataset enables robust cross-lingual learning and allows one multilingual checkpoint to serve all six languages at the cost of additional compute (Table~\ref{tab:compute_cost}), whereas training one language at a time is less compute-intensive but can lead to catastrophic forgetting when adapting the model to another language.

\subsection{Domain Robustness} 
\label{sec:domain_robustness}
\begin{figure*}[!t]
\centering
\begin{subfigure}{0.49\textwidth}
\centering
\includegraphics[width=\linewidth]{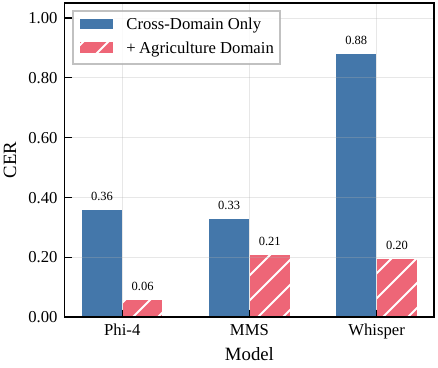}
\caption{In-domain (held-out agricultural split) \protect\footref{fn:dg_green_kikuyu}.}
\label{fig:domain_robustness}
\end{subfigure}\hfill
\begin{subfigure}{0.49\textwidth}
\centering
\includegraphics[width=\linewidth]{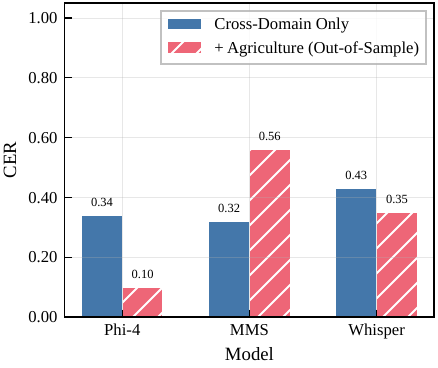}
\caption{Out-of-sample agricultural test set \protect\footref{fn:kik_private}.}
\label{fig:domain_robustness_oos}
\end{subfigure}
\caption{Domain robustness (CER) for Kikuyu with and without in-domain agricultural fine-tuning.}
\label{fig:domain_robustness_all}
\end{figure*}

We report domain robustness for Kikuyu (\texttt{kik}), the only language with public agricultural-domain data, under the three conditions defined in Section~\ref{sec:domain_adaptability}. As shown in Figure~\ref{fig:domain_robustness} and Figure~\ref{fig:domain_robustness_oos}, \textbf{domain adaptation is high-reward but architecture-sensitive.} Phi-4 achieves up to 83\% relative CER reduction in-domain and Whisper up to 70\% out-of-sample (Figures~\ref{fig:domain_robustness} and~\ref{fig:domain_robustness_oos}), demonstrating adaptation and generalization to unseen agricultural data.
In contrast, MMS improves in-domain (0.33 to 0.21 CER) but degrades out-of-sample (0.32 to 0.56), suggesting that its lightweight adapter absorbs domain-specific patterns without robust transfer. Small amounts of domain data can therefore benefit LLM-based and encoder--decoder models, while adapter-based models require careful monitoring for out-of-domain degradation.

\subsection{New versus Proxy MMS Adapters}
\label{sec:adapter_strategy}
\begin{figure}[!t]
\centering
\includegraphics[width=\columnwidth]{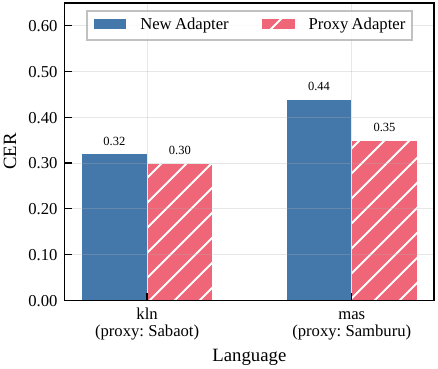}
\caption{MMS adapter strategies (CER): training new language-specific adapters versus reusing related-language proxy adapters for Kalenjin and Maasai. Corresponding WER is reported in Figure~\ref{fig:mms_adapter_strategy} (Appendix~\ref{sec:appendix_wer}).}
\label{fig:mms_adapter_strategy_cer}
\end{figure}

As shown in Figures~\ref{fig:mms_adapter_strategy_cer} and~\ref{fig:mms_adapter_strategy}, \textbf{related-language initialization outperforms training from scratch.} Fine-tuning MMS adapters using related proxy languages yields small but consistent gains of 6--20\% CER improvement and 3--9\% WER improvement over training new adapters from scratch. The newly trained adapters achieve reasonable performance, though consistently lower than the proxy adapters given the same fine-tuning data. These results indicate that leveraging language-family proximity where possible provides a stronger initialization for MMS adaptation, yielding more effective parameter reuse and better generalization than adapters trained from scratch where data is minimal. These results may differ with more data for new language adapter training.

\subsection{Effects of Typological Complexity}
\label{sec:effect_language_complexity}
\begin{figure}[!t]
\centering
\includegraphics[width=\columnwidth]{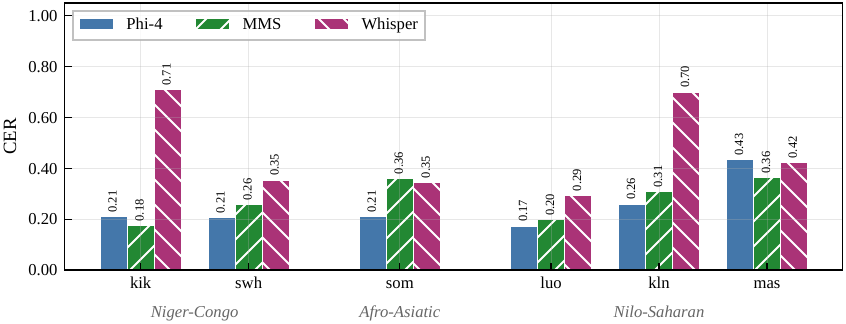}
\caption{CER by language grouped by family (Niger-Congo, Nilo-Saharan, Afro-Asiatic) across all three models at 200\,h, showing the correlation between typological complexity and error rate. Corresponding WER is reported in Figure~\ref{fig:language_complexity} (Appendix~\ref{sec:appendix_wer}).}
\label{fig:language_complexity_cer}
\end{figure}

The 200\,h monolingual CER results in Figures~\ref{fig:model_comparison_cer} and~\ref{fig:language_complexity_cer} reveal a striking stratification of model performance that correlates with language family and typological complexity. Niger-Congo languages Swahili and Kikuyu consistently achieve the lowest CER across most models. Dholuo, a Nilo-Saharan language with relatively transparent phonology, achieves comparably low CER. In contrast, the Nilo-Saharan languages Maasai and Kalenjin, characterized by complex vowel harmony, tonal contrasts, and highly agglutinative morphology, consistently yield the highest CER across all three architectures.

Somali (\texttt{som}, Afro-Asiatic, Cushitic) occupies an intermediate position, reflecting its phonological characteristics, a moderate consonantal inventory and non-tonal structure, which may have influenced pre-training data coverage for Whisper \citep{radford2023whisper}.

The elevated CER for Whisper on Kikuyu (0.71) and Kalenjin (0.70) compared to Phi-4 (0.21 and 0.26, respectively) further suggests that character-level accuracy is particularly sensitive to tokenizer design for morphologically rich languages. Whisper's subword tokenizer, optimized for high-resource languages, may produce systematically incorrect segmentations for agglutinative word forms, compounding errors from misaligned token boundaries \citep{LiuYangQu2024WhisperFinetuning}.

\subsection{Training Cost and Resource Efficiency}
\label{sec:training_cost_efficiency}
Table~\ref{tab:compute_cost} compares the training cost and resource requirements of the three models. We report results for the 200-hour baseline fine-tuning checkpoint. Phi-4 and Whisper are trained in a multilingual setup, whereas MMS leverages language-specific adapters that can be fine-tuned independently and in parallel. For monolingual fine-tuning, the reported training time corresponds to the average wall-clock time across independent fine-tuning jobs for the six languages.
\textbf{Accuracy and resource efficiency involve different trade-offs} as seen in Section~\ref{sec:hardware}). We therefore report peak GPU memory usage and wall-clock training time as empirical measures to characterize the practical cost of training each model, rather than as differences in fine-tuning strategy.

\begin{table}[t!]
\centering
\small
\setlength{\tabcolsep}{6pt}
\begin{tabular}{l c c c}
\toprule
\textbf{Model} & \textbf{VRAM} & \multicolumn{2}{c}{\textbf{Train Time (200\,h)}} \\
\cmidrule(lr){3-4}
 & \textbf{(GB)} & \textbf{Multi} & \textbf{Mono} \\
\midrule
Whisper & 29.1 & 79  & 26 \\
Phi-4   & 71.2 & 53  & 6  \\
MMS     & 26.6 & N/A & 15 \\
\bottomrule
\end{tabular}
\caption{Peak GPU VRAM usage and training time per model for baseline fine-tuning (hours)}
\label{tab:compute_cost}
\end{table}

\section{Conclusion}
We presented a comprehensive empirical evaluation of fine-tuning strategies for ASR in five low-resource African languages and Swahili as a mid-resource African language, spanning monolingual and multilingual adaptation, data scaling, domain robustness, and language-adapter transfer. Our main conclusions are: (1) Phi-4-multimodal-instruct achieves the lowest CER on four of six languages, demonstrating that LLM based multimodal architectures transfer effectively to low-resource settings even without explicit language coverage; (2) data sufficiency is model-dependent: MMS saturates by 150\,h (paired $t$-test, $p{=}0.036$) while Phi-4 continues improving through 300\,h on most languages, establishing practical data collection thresholds; (3) domain-specific fine-tuning yields dramatic gains for Phi-4 and Whisper but causes MMS to degrade out-of-sample;  (4) proxy adapters from linguistically related languages consistently outperform new adapter initialization, with a WER improvement of up to 9\% relative for Maasai; and  (5) language family and typology impacts model performance.

These findings provide actionable guidance for researchers and developers building speech technologies for African languages to invest in high-capacity architectures, prioritize domain-specific data collection over volume scaling beyond 300 hours, account for typological complexity in model adaptation and evaluation choices and use linguistic proximity for transfer decisions. We release all training configurations to support reproducibility and future work on African language ASR.

\section*{Limitations}
Our finetuning evaluation spans three ASR architectures, one pretrained model each. The results reflect model-level differences rather than architectural causation. Models differ in pre-training data, language coverage, tokenizer, parameter count and trainable components which can confound architecture comparisons. Additional models per architecture are needed to further validate our findings. 

Typological analysis is correlational, not causal. Six languages, without quantitative typological features or a model linking them to ASR error, cannot establish language complexity as the cause of performance differences. Corpus composition, dialects, recording conditions, and speech style may contribute: Swahili comes from Common Voice, while other languages mostly come from ANV-KE. Wider corpus coverage and quantitative typological modeling are needed. 

Despite matched per-language budgets, joint training across all languages exposes multilingual models to more total speech, so performance differences cannot be attributed solely to multilingual versus monolingual fine-tuning. 

Data scaling suggests diminishing returns under tested conditions, not definitive saturation or universal sufficiency thresholds. Paired t-tests across six languages and unrepeated, single-seed training limit variance estimates and plateauing claims. Significance differs from practical effect size: MMS’s small change from 150 h to 450 h is significant, but Phi-4’s larger average CER improvement is not. Non-significance does not imply no benefit from more data.

Kikuyu domain adaptation adds 38.3 h of agricultural speech to approximately 200 h of general-domain speech, confounding domain composition with training quantity. Without matched general-domain-data control, domain-scaling evidence is indicative, not conclusive, about domain-match versus additional-data effects.

\section*{Ethical Considerations}
All speech data used were obtained under appropriate usage terms and contain no personally sensitive data. The datasets may not fully capture the dialectal, speaker, and recording-condition diversity of each language, so reported error rates need not generalize to all speakers, dialects or regions where a language is spoken; deployment in high-stakes settings should be preceded by community-representative evaluation. Because performance varies across languages, remaining poor for the Nilo-saharan languages, practitioners should not assume parity of service across the six languages.

\clearpage

\bibliography{anthology}

\clearpage

\appendix
\section{Appendix}
\label{sec:appendix_wer}
This appendix reports the Word Error Rate (WER) companions to the CER figures in the main body, the full data-sufficiency analysis table and fine-tuning hyperparameters used in all experiments. Consistent with our focus on CER for morphologically rich low-resource languages, WER trends mirror the CER findings.

\begin{figure}[h]
\centering
\includegraphics[width=\columnwidth]{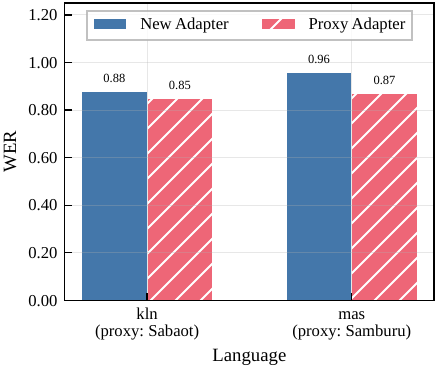}
\caption{MMS adapter strategies (WER): training new language-specific adapters versus reusing related-language proxy adapters for Kalenjin and Maasai (companion to Figure~\ref{fig:mms_adapter_strategy_cer}).}
\label{fig:mms_adapter_strategy}
\end{figure}

\begin{figure}[h]
\centering
\includegraphics[width=\columnwidth]{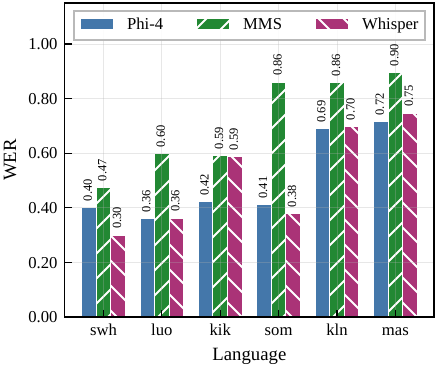}
\caption{WER across six languages for Phi-4, MMS, and Whisper fine-tuned on 200\,h of monolingual data (companion to Figure~\ref{fig:model_comparison_cer}).}
\label{fig:model_comparison_wer}
\end{figure}

\begin{figure}[h]
\centering
\includegraphics[width=\columnwidth]{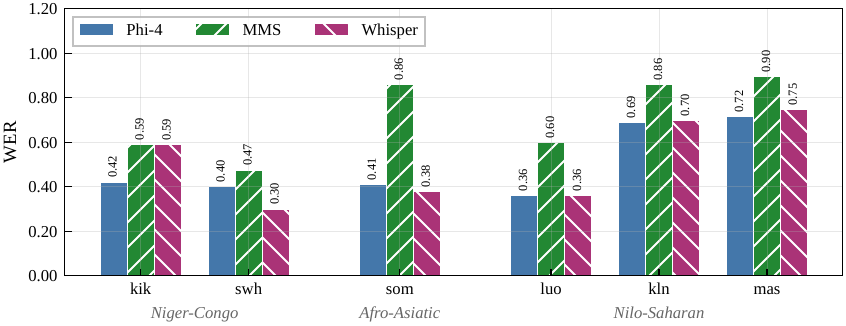}
\caption{WER by language grouped by family across all three models at 200\,h (companion to Figure~\ref{fig:language_complexity_cer}).}
\label{fig:language_complexity}
\end{figure}

\begin{table}[t]
  \centering
  \setlength{\tabcolsep}{2.5pt}
  \small
  \begin{tabular}{@{}llrrr@{}}
    \toprule
    \textbf{Model} & \textbf{Lang.} & \textbf{CER\textsubscript{150h}} & \textbf{CER\textsubscript{450h}} & \textbf{$\Delta$ CER} \\
    \midrule
    MMS & kik & 0.18 & 0.17 & $-$0.01 \\
     & kln & 0.31 & 0.30 & $-$0.01 \\
     & luo & 0.19 & 0.18 & $-$0.01 \\
     & mas & 0.36 & 0.36 & \phantom{$-$}0.00 \\
     & som & 0.37 & 0.36 & $-$0.01 \\
     & swh & 0.26 & 0.25 & $-$0.01 \\
    \cmidrule(l){2-5}
    \multicolumn{2}{@{}l}{\textit{Mean $\Delta$ (pp)}} & \multicolumn{3}{r}{$-$0.7}  \\
    \multicolumn{2}{@{}l}{\textit{Paired $t$-test}} & \multicolumn{3}{r}{$t{=}2.83,\quad p{=}0.036${*}} \\
    \multicolumn{2}{@{}l}{\textit{Bootstrap 95\,\% CI (pp)}} & \multicolumn{3}{r}{$[-1.0,\;-0.5]$} \\
    \midrule
    Phi-4 & kik & 0.35 & 0.33 & $-$0.02 \\
     & kln & 0.35 & 0.26 & $-$0.09 \\
     & luo & 0.26 & 0.17 & $-$0.09 \\
     & mas & 0.58 & 0.48 & $-$0.10 \\
     & som & 0.26 & 0.24 & $-$0.02 \\
     & swh & 0.21 & 0.26 & +0.05 \\
    \cmidrule(l){2-5}
    \multicolumn{2}{@{}l}{\textit{Mean $\Delta$ (pp)}} & \multicolumn{3}{r}{$-$4.4}  \\
    \multicolumn{2}{@{}l}{\textit{Paired $t$-test}} & \multicolumn{3}{r}{$t{=}1.87,\quad p{=}0.13$} \\
    \multicolumn{2}{@{}l}{\textit{Bootstrap 95\,\% CI (pp)}} & \multicolumn{3}{r}{$[-8.3,\;+0.2]$} \\
    \bottomrule
    \multicolumn{5}{@{}l}{\scriptsize {*}\,$p<0.05$. Paired two-sided $t$-test over six per-language}\\
    \multicolumn{5}{@{}l}{\scriptsize CER differences; bootstrap CI: 10\,000 resamples.}
  \end{tabular}
    \caption{Data sufficiency analysis comparing CER at 150\,h and the final checkpoint (450\,h) for MMS and Phi-4 across six languages. Paired significance tests are conducted per model ($n{=}6$ languages). Negative $\Delta$ denotes improvement and positive $\Delta$ denotes degradation.}
  \label{tab:pct_change}
\end{table}

\begin{table}[t]
\centering
\small
\begin{tabular}{lccc}
\toprule
\textbf{Model} &
\shortstack{\textbf{Effective}\\\textbf{Batch Size}} &
\shortstack{\textbf{Learning}\\\textbf{Rate}} &
\shortstack{\textbf{Weight}\\\textbf{Decay}} \\
\midrule
Whisper  & 256 & $1\times10^{-5}$ & 0.01 \\
Phi-4  & 96 & $1\times10^{-4}$ & 0.005 \\
MMS  & 128 & $1\times10^{-3}$ & 0.01 \\
\bottomrule
\end{tabular}
\caption{Fine-tuning hyperparameters used in all experiments. All experiments use the AdamW optimizer.}
\label{tab:training_hparams}
\end{table}

\end{document}